\documentclass[runningheads]{llncs}
\usepackage{graphicx}
\usepackage{comment}
\usepackage{amsmath,amssymb} 
\usepackage{color}
\let\vec\mathbf
\usepackage{enumitem}
\DeclareMathOperator*{\argmax}{arg\,max}
\usepackage{xcolor,soul}
\sethlcolor{lightgray}
\usepackage{multicol}
\usepackage{hyperref}
\usepackage{caption}
\usepackage{subcaption}

\begin{document}
\pagestyle{headings}
\mainmatter
\def\ECCVSubNumber{3056}  

\title{Knowledge-Based Video Question Answering with Unsupervised Scene Descriptions} 

\titlerunning{Knowledge-Based VideoQA with Unsupervised Scene Descriptions}
%
\author{Noa Garcia \and Yuta Nakashima}
\authorrunning{Garcia and Nakashima}
%
\institute{Osaka University, Japan \\
\email{\{noagarcia, n-yuta\}@ids.osaka-u.ac.jp}}
\maketitle

\begin{abstract}
To understand movies, humans constantly reason over the dialogues and actions shown in specific scenes and relate them to the overall storyline already seen. Inspired by this behaviour, we design ROLL, a model for knowledge-based video story question answering that leverages three crucial aspects of movie understanding: dialog comprehension, scene reasoning, and storyline recalling. In ROLL, each of these tasks is in charge of extracting rich and diverse information by 1) processing scene dialogues, 2) generating unsupervised video  scene descriptions, and 3) obtaining external knowledge in a weakly supervised fashion. To answer a given question correctly, the information generated by each inspired-cognitive task is encoded via Transformers and fused through a modality weighting mechanism, which balances the information from the different sources. Exhaustive evaluation demonstrates the effectiveness of our approach, which yields a new state-of-the-art on two challenging video question answering datasets: KnowIT VQA and TVQA+.
\end{abstract}

\section{Introduction}

Robots may not dream of electric sheep yet,\footnote{`Do androids dream of electric sheep?' (Philip K. Dick, 1968).} but in the last few years, artificial intelligence has shown significant progress towards human-like reasoning. This has been made possible by emulating snippets of human intelligence in constrained tasks \cite{antol2015vqa,johnson2017clevr}, where machine performance is easily evaluated. Among those tasks, video story question answering \cite{tapaswi2016movieqa,lei2018tvqa,garcia2020knowit} emerged as a testbed to approximate real-world situations, in which not only the spatial relationships between objects are important, but also the temporal coherence between past, present, and future events.

\begin{figure}
\centering
\includegraphics[width = \textwidth]{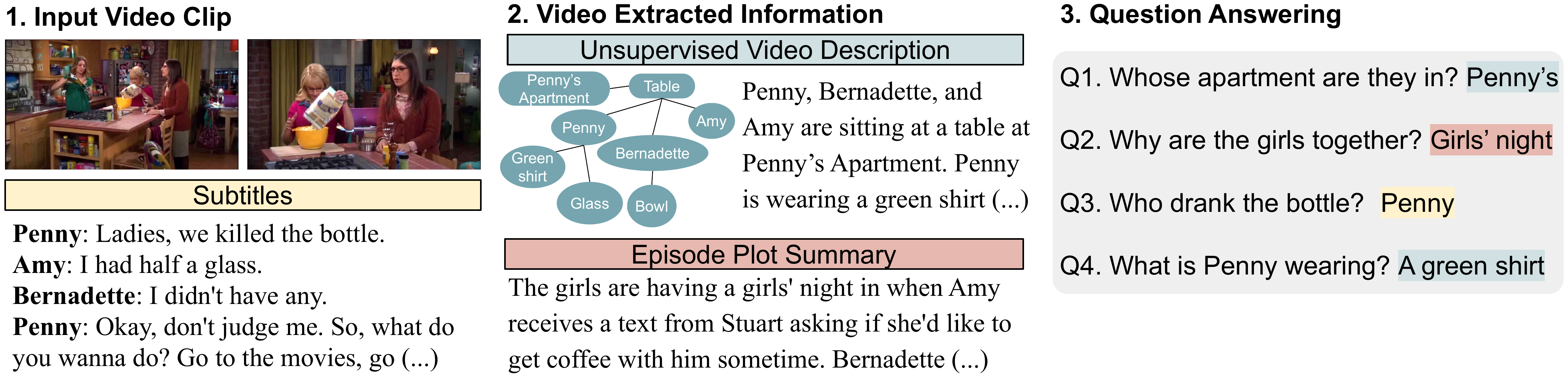}
\caption{ROLL performs video story question answering by generating unsupervised descriptions from video scene graphs and obtaining episode summaries.} 
\label{fig:into}
\end{figure}

Video story question answering leverages the structure of video stories, such as movies and TV shows, to formulate questions about specific scenes in a video. Models, then, need to find the correct answer by reasoning over the scene and its underlying plot. However, as the video story unfolds, the details of the plot are often revealed to the spectator over multiple scenes, sometimes far apart from each other. To understand the whole story, humans have the capacity to constantly relate past events with what is currently being shown, acquiring contextual information that forms their story knowledge. We argue that for a full comprehension of video stories, not only what is happening in the current scene has to be considered, but also the knowledge acquired in previous scenes. Some examples are shown in Fig.~\ref{fig:into}; whereas the answer to Q1, Q3, and Q4 can be guessed from the video scene (and its subtitles), Q2 can only be inferred when the full context is known.

Previous work on video story question answering can be roughly divided into two categories. On one hand, there are models that extract information from the whole video story \cite{tapaswi2016movieqa,na2017read,liang2018focal,kim2018multimodal}, and use attention mechanisms to find the parts that are relevant to each question. These models obtain contextual representations, which are used to answer general questions about the plot, but barely capture details at the scene level. On the other hand, other models extract detailed information from specific scenes \cite{lei2018tvqa,kim2019progressive}, without looking at the whole video story. However, relying only on the content of short scenes is insufficient to answer insightful aspects about the story, such as the characters' motivations. To study multiple types of questions about video stories using both contextual and scene-specific information, a knowledge-based video question answering dataset has been recently introduced \cite{garcia2020knowit}. The proposed model combines contextual information from external resources with multi-modal representations from specific scenes. However, the contextual data in \cite{garcia2020knowit} is obtained from thousands of task-specific human-generated annotations, which are expensive to obtain and difficult to generalise to other domains.

In this paper, we introduce ROLL, \underline{R}ead, \underline{O}bserve, and Reca\underline{ll}, a model that addresses knowledge-based video story question answering with both contextual and scene-specific information using unsupervised scene descriptions and weakly-supervised external knowledge. ROLL consists on a three-branch architecture inspired by three areas of human cognition playing an important role in video understanding: dialog comprehension (\textit{read branch}), scene reasoning (\textit{observe branch}), and storyline recalling (\textit{recall branch}). Whereas the scene-specific details are summarised in the read and observe branches, the recall branch provides a contextual overview about the story using free online resources. To predict the correct answer, the three branches are lately fused through a  modality weighting mechanism, which balances the signal from the three different sources. 

\noindent
\textbf{Contributions:} Our contribution is three-fold: 1)~we propose a new unsupervised video representation based on video descriptions generated from video scene graphs; 2)~we combine specific details from video scenes with weakly supervised external knowledge for a deep understanding of video stories; and 3)~we incorporate a modality weighting mechanism to fuse data from different modalities without information loss. Our model is evaluated on two challenging video story question answering datasets: KnowIT VQA and TVQA+, outperforming previous work by more than 6.3\% and 1.3\%, respectively.
\section{Related Work}
\label{sec:relatedwork}

We develop a model for video story question answering that 1) takes advantage of rich external knowledge sources, and 2) represents video content by generating unsupervised video captions from scene graphs. In the following, we first review work on video story question answering and visual reasoning with external knowledge before discussing scene graphs and methods for video description.

\noindent
\textbf{Video Story Question Answering} Video story question answering is a modality in video question answering in which  questions are not only related to the visual content of a video, but also to its plot. MovieQA \cite{tapaswi2016movieqa} introduced a plot-oriented dataset with questions generated from movie summaries. Most proposed models \cite{tapaswi2016movieqa,na2017read,liang2018focal,kim2018multimodal} used frame-level features to represent the entire movie, applying attention mechanisms to find the relevant parts to each question. This provides a high-level overview of the story, but does not consider the details of each scene. Alternatively, PororoQA \cite{kim2017deepstory} and TVQA \cite{lei2018tvqa} formulated scene-level questions about specific events in the video. Models addressing these datasets described the details of each scene with features \cite{wang2018movie}, captions \cite{kim2017deepstory} or visual concepts \cite{lei2018tvqa,kim2019progressive,yang2020bert}, but without attending to the ongoing plot in the video story. Recently, KnowIT VQA \cite{garcia2020knowit} introduced a combination of detailed questions about scenes and knowledge-based questions about the story. The proposed model relied on human-generated annotations to understand the insights of the plot. On the contrary, our model exploits both specific and general story information without task-specific annotations by using external knowledge bases.

\noindent
\textbf{Visual Reasoning with External Knowledge} Using external knowledge in visual reasoning extends the visual question answering task (VQA) to address questions far beyond the visual content of images. Although the acquisition of knowledge depends on the task of interest, structured knowledge bases, such as DBpedia \cite{auer2007dbpedia} or ConceptNet \cite{speer2017conceptnet}, are commonly used in most methods \cite{wu2016ask,wang2017explicit,wang2018fvqa,narasimhan2018straight,narasimhan2018out}. However, structured knowledge is usually represented as (subject, predicate, object) triplets, which is a hard constraint on the type of information being processed. Generic solutions \cite{shah2019kvqa,marino2019ok} proposed to exploit unstructured resources in natural language, such as Wikipedia.\footnote{\href{https://www.wikipedia.org/}{https://www.wikipedia.org/}} Following this direction, our model leverages unstructured online data to answer knowledge-based questions about video stories.

\noindent
\textbf{Scene Graphs} Scene graphs \cite{johnson2015image} are structures that represent the objects depicted in an image and their relationships, providing a semantic description of the image. Most scene graph methods consist on an object detector, an attribute classifier and a relationship predictor \cite{xu2017scene,li2017scene,zellers2018neural,yang2018graph,zhang2019large,zhang2019graphical}. Scene graphs have been used in multiple vision and language tasks, including image captioning \cite{yang2019auto,Gu_2019_ICCV,zellers2018neural} and VQA \cite{teney2017graph,shi2019explainable,xiong2019visual}. However, less attention has been paid to generating scene graphs from videos, in which relationships are both spatial and temporal. So far, video scene graphs have been mostly applied to cross-modal retrieval to find video fragments \cite{xiong2019graph,vicol2018moviegraphs}. In this work, we rely on video scene graphs to generate unsupervised video scene descriptions.

\noindent
\textbf{Video Descriptions} Video captioning aims to describe short video clips using natural language. Most approaches \cite{yao2015describing,pan2016jointly,chen2018less,liu2018sibnet,wang2019controllable} use a sequential encoder-decoder framework, in which the input are visual features from multiple frames and the output is the generated sentence. For more detailed descriptions, dense video captioning \cite{krishna2017dense,zhou2018end} generates multiple sentences describing all the relevant events in the video. However, existing methods require to be trained on large-scale annotated datasets \cite{rohrbach2015dataset,pini2019m} with thousands of video-description pairs. We generate rich video scene descriptions in an unsupervised way using the semantic information from video scene graphs.

\section{Model Overview}
\begin{figure}[t]
\centering
\includegraphics[width = 0.95\textwidth]{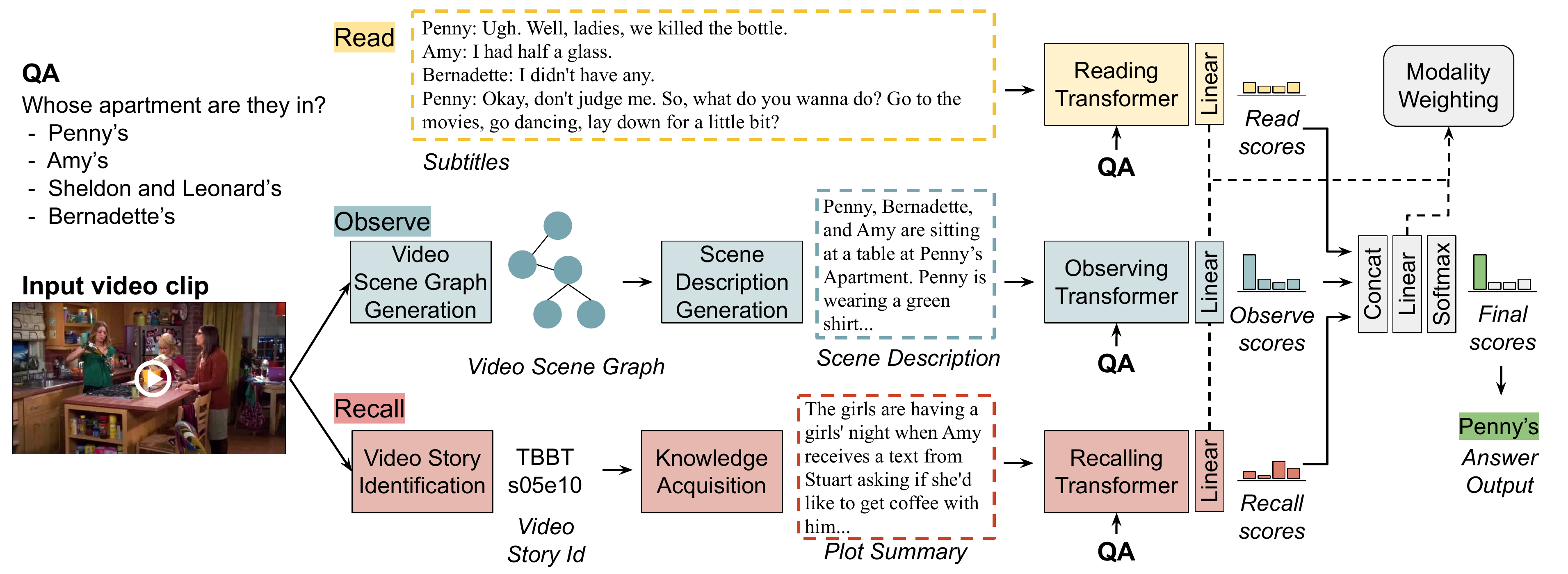}
\caption{ROLL overview. Each branch estimates a relevance score for each of the candidate answers based on different information. The read branch relies on subtitles, the observe branch generates unsupervised video descriptions, and the recall branch obtains external knowledge as plot summaries. To predict the correct answer, the three outputs are fused through a modality weighting mechanism.} 
\label{fig:model}
\end{figure}

The goal of video story question answering is to understand movies or TV shows in a similar way as we humans do. We argue that there are at least three aspects of human intelligence involved in this task: 1) comprehension of what is being said, 2) comprehension of about what is being watched, and 3) recalling what happened in the story before. Our proposed model, ROLL, emulates each of those aspects in a three branch architecture, as shown in Fig.~\ref{fig:model}. Each branch in ROLL (read, observe, and recall) represents as text data the information from a different cognitive task, and encodes it through a Transformer with several self-attention layers. Then, the outputs from each Transformer are fused through a modality weighting mechanism to predict the correct answer. 

\noindent
\textbf{Task definition} We address video story question answering as a knowledge-based multiple-choice task. For each sample, the input is: 1) a question, 2) $N_{ca}$ candidate answers, 3) a video scene, and 4) the subtitles associated with the scene. The output is the index of the predicted answer. As a knowledge-based task, models can access external resources to retrieve contextual information.

\noindent
\textbf{Introduction to Transformers} Transformers \cite{vaswani2017attention} are sequence-to-sequence modelling architectures that entirely rely on self-attention mechanisms. They have rapidly become the state-of-the-art in many natural language processing tasks. ROLL incorporates three independent Transformers to model the language data extracted from each branch, which is represented by the input string: %
\begin{equation}
\footnotesize
    {s_m}^c = \text{[CLS]} + \text{context}_m +  \text{[SEP]} + {\text{choice}_m}^c +  \text{[SEP]},
    \label{eq:stringinput}
\end{equation} %
where $m$ indicates the branch, $\small \text{context}_m$ is an input sentence defined for each branch, $\small {\text{choice}_m}^c$ is a sentence for the $c$-th candidate answer with $c=1,\cdots,N_{ca}$, {\small [CLS]} is the classification token used to obtain the output representation, {\small [SEP]} is the separator token for differentiating  sentences, and $+$ is string concatenation. For each sample, $N_{ca}$ input strings are generated, one per candidate answer.

The input string ${s_m}^c$ is tokenised into a sequence of $n$ tokens $\vec{x}^c = [x_1, \cdots, x_n]$, and fed into a Transformer network. For each token $x_i$ in $\vec{x}^c$, the Transformer creates an input embedding, $\vec{h}_i^0 \in \mathbb{R}^{D_h}$ with $D_h$ hidden size, by adding the word, segment, and position embeddings. For each self-attention layer $l = 1, \cdots, N_L$ in the Transformer, denoted by $\text{TBlock}^l(\cdot)$, the contextualised word representation for position $i$ in the sequence is computed as: %
\begin{equation}
    \vec{h}_i^l = \text{TBlock}^l(\vec{h}_i^{l-1})
\end{equation} %
The encoded representation of the input string ${s_m}^c$ is then obtained as the output of the position of the {\small [CLS]} token in the last layer: %
\begin{equation}
    {\vec{y}_m}^c = \vec{h}_0^{L} \in \mathbb{R}^{D_h}
\end{equation} %
Our Transformers are the ${\text{BERT}_{\text{BASE}}}$ model~\cite{devlin2019bert} with ${N_L = 12}$ and ${D_h = 768}$.


\section{Read Branch}
In the read branch, ROLL extracts information from the dialogues of the video scene, which are obtained from the subtitles. The input string for this branch is: %
\begin{equation}
    {s_r}^{c} = \text{[CLS]} + subs + q + \text{[SEP]} + a^c + \text{[SEP]},
\end{equation} %
where $subs$ are the subtitles, $q$ the question, and $a^c$ with ${c=1,\cdots,N_{ca}}$ each of the candidate answers. Each input string ${s_r}^c$ is fed into the Reading Transformer to obtain ${\vec{y}_r}^c$, which is forwarded into a single output linear layer with ${\vec{w}_r}$ weights and $b_r$ bias, to compute a \textit{read score} per candidate answer: %
\begin{equation}
{\alpha_r}^c = {\vec{w}_r}^\top \cdot {\vec{y}_{r}}^c + b_r
\end{equation}


\section{Observe Branch}
In the observe branch, ROLL summarises the semantics of the video scene into a video description. Generating descriptions from video is a challenging problem~\cite{xu2016msr}. Standard video captioning models \cite{chen2018less,liu2018sibnet,wang2019controllable} require to be trained on large-scale datasets with annotated video and description pairs. As video story question answering datasets commonly do not provide such annotations, training a model for our task is impractical. Similarly, relying on pre-trained models may lead to poor results, as the generated descriptions will miss important information about the story (e.g., character names or frequent locations). Alternatively, we propose to generate unsupervised video descriptions by first creating a video scene graph. The descriptions are then fed into the Observing Transformer to predict a \textit{observe score} for each candidate answer. Below, we first describe the video scene graph generation process, then we provide the details for the unsupervised video description, and finally we summarise the observing Transformer.

\subsection{Video Scene Graph Generation}
\begin{figure}[t]
\centering
\begin{minipage}{.47\textwidth}
  \centering
\includegraphics[width = 0.98\textwidth]{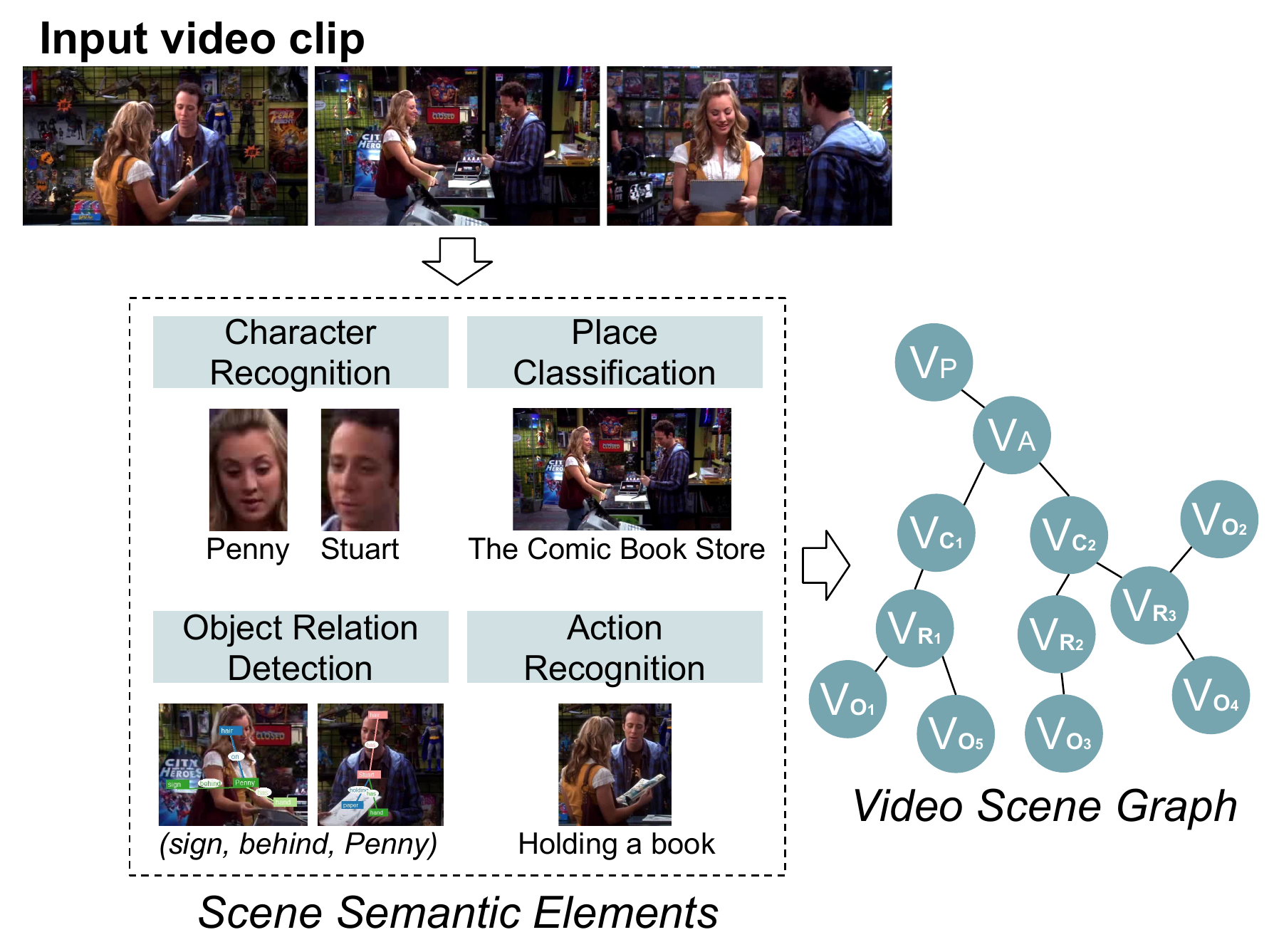}
  \captionof{figure}{Video scene graphs are generated from recognising the semantic elements of the scene (characters $V_C$, places $V_P$, objects $V_O$, relations $V_R$, and actions $V_A$) and connecting them.}
  \label{fig:scenegraph}
\end{minipage} \hspace{10pt}
\begin{minipage}{.47\textwidth}
  \centering
\includegraphics[width = 0.93\textwidth]{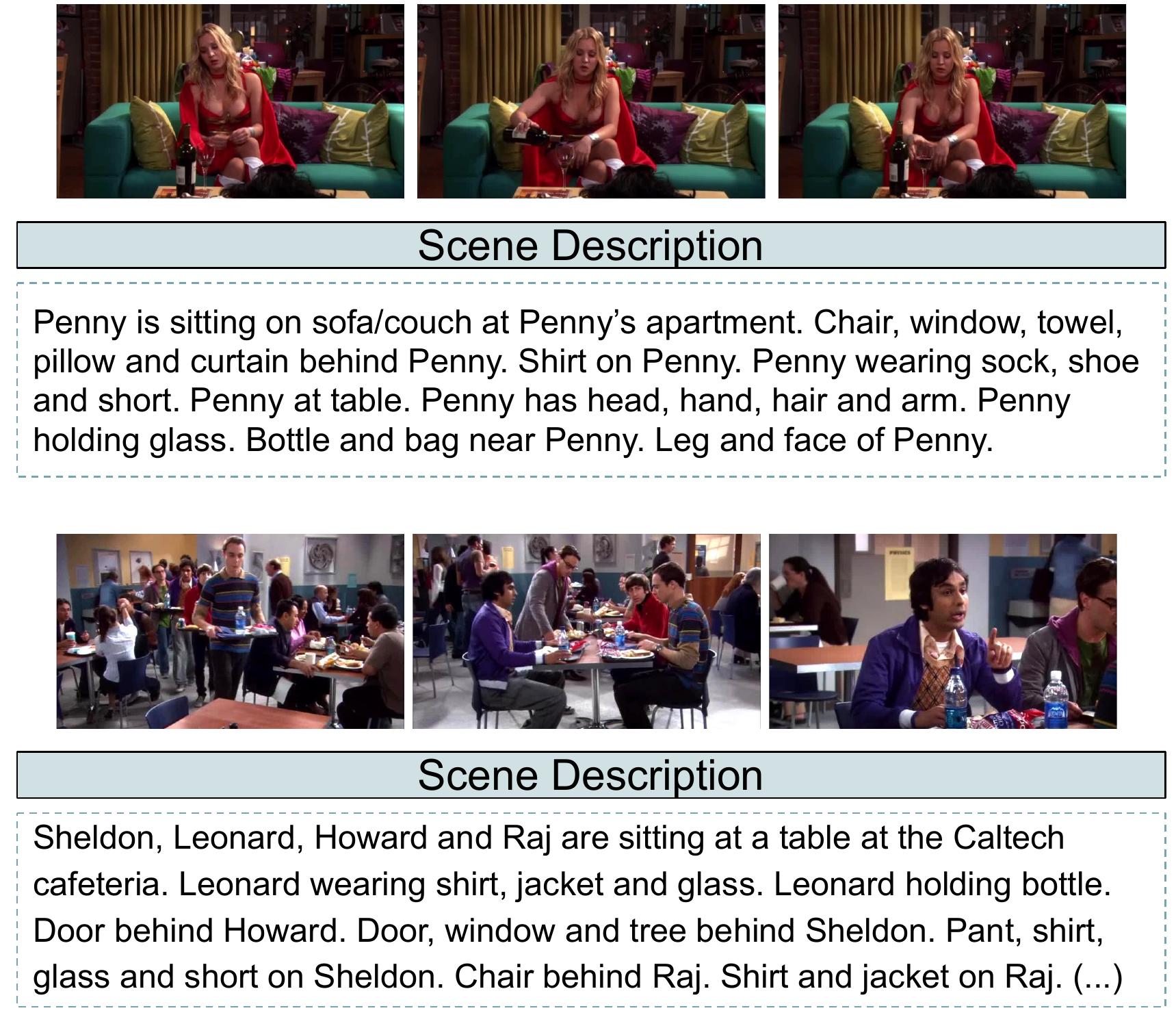}
  \captionof{figure}{Examples of generated scene description. Although not natural, they accurately represents the semantics in the video scene.}
  \label{fig:description}
\end{minipage}
\end{figure}

Fig.~\ref{fig:scenegraph} shows the video scene graph generation process, which is built on top of state-of-the-art image and video recognition techniques. We use four modules to detect the most relevant details in the scene: character recognition, place classification, object relation detection, and action recognition. The video scene graph is then generated by building connections between the detected elements.

\noindent
\textbf{Character Recognition}
This module identifies the characters that appear in the scene using a face recognition classifier trained with images from the cast. We download about 10 images for each of the most common $N_{C_T}$ characters based on IMDb.\footnote{\href{https://www.imdb.com/}{https://www.imdb.com/}} We extract $f \in \mathbb{R}^{128}$ face representations with FaceNet \cite{schroff2015facenet} and train a $k$-nearest neighbour (kNN) classifier, where $k = N_{C_T}$. At test time, the trained kNN classifier returns a score for the predicted character. If the score is below a threshold, we assigned it to the \textit{unknown} class. Finally, we apply a spatio-temporal filter to remove mispredictions and duplicate characters. Details are provided in the suppl. material. As output, we obtain a set of $N_C$ characters appearing in the scene $C = \{C_i | i = 1, \cdots, N_C\}$, and their bounding boxes.
    
\noindent
\textbf{Place Classification} The place classification module detects where the scene is located. To learn the frequent locations in the video story, we fine-tune the pre-trained Places365~\cite{zhou2017places} network with ResNet50~\cite{he2016deep} backbone in a weakly supervised way. To obtain place annotations, we use video transcripts from specialised websites.\footnote{For example, \href{https://bigbangtrans.wordpress.com/}{https://bigbangtrans.wordpress.com/}.} We extract the locations that appear at least 10 times in the training set scripts, and include an \textit{unknown} category for the rest. Training is performed at the frame level, i.e., each frame is considered as an independent image. For prediction, we accumulate the scores of the top 5 predicted classes for each frame in a video scene and output the most scored place, $P$.
    
\noindent
\textbf{Object Relation Detection} This module detects the objects in the scene and their relations. We use the large-scale visual relationship understanding (VRU) \cite{zhang2019large} pre-trained on the VG200 dataset \cite{xu2017scene}, with 150 object and 50 relation categories. For each frame, VRU returns a list of subject-relation-object triplets, their bounding boxes, and a prediction score for each triplet. We replace the objects and subjects assigned to a person class\footnote{Boy, girl, guy, lady, man, person, player, woman.} with its corresponding character name by finding the overlap between the bounding boxes. We only keep triplets assigned to known characters and we filter out duplicates. After discarding the bounding boxes and scores, we obtain a list of $N_T$ triplets, $T = \{T_i | i = 1, \cdots, N_T\}$ with $T_i = (S_i, R_i, O_i)$ and $S_i$, $R_i$, and $O_i$ as  subject, relation, and object.

\noindent
\textbf{Action Recognition} The action recognition module detects the main action in the video scene. We use Long-Term Feature Banks (LFB) \cite{wu2019long} pre-trained on the Charades dataset \cite{sigurdsson2016hollywood} with 157 action categories. LFB extracts information over the entire span of the video scene, improving performance with respect to using short 2-3 second clips. We input the entire scene into the network, and we obtain a predicted action as a result, $A$.

\noindent
\textbf{Graph Generation}
The video scene graph, $G=(V,E)$, semantically describes the visual contents of the scene by using a collection of nodes $V$, and edges $E$. We consider the following types of nodes: %
\begin{itemize}[noitemsep,topsep=0pt]
\item[$\bullet$] \textit{Character nodes}, $V_C \subseteq V$, representing the characters in the scene. If $C$ do not contain any \textit{unknown} character, $V_C = C$. Otherwise, we remove the \textit{unknown} characters $\{\textsc{unk}_C\}, $ as $V_c = C - \{\textsc{unk}_C\}$.

\item[$\bullet$] \textit{Place nodes}, $V_P \subseteq V$, representing the location where the video scene occurs. $V_P = \{P\}$ if $P \neq \textit{unknown}$, otherwise $V_P = \emptyset$.

\item[$\bullet$] \textit{Object nodes}, $V_O \subseteq V$, representing the objects in the scene, which are obtained from the subjects and objects in the triplets that are not a character, as ${V_O = Z - ( Z \cap C )}$ with ${Z = S \cup O}$. 

\item[$\bullet$] \textit{Relation nodes}, $V_R \subseteq V$, representing the relation between subjects and objects in the triplets, $V_R = R$. 

\item[$\bullet$] \textit{Action nodes}, $V_A \subseteq V$, representing the action in the scene as $V_A = \{A\}$, with $|V_A| = 1$.
\end{itemize} %
We use 6 types of directed edges:%
\begin{itemize}[noitemsep,topsep=0pt]
\item[$\bullet$] $e_{P,A}=(V_{P},V_{A}) \in E$ between the place node $V_{P}$ and the action node $V_{A}$.
\item[$\bullet$] $e_{A,C_j}=(V_{A},V_{C_j}) \in E$ between the action node $V_{A}$ and each character  $V_{C_j}$.
\item[$\bullet$] ${e_{C_i,R_j}=(V_{C_i},V_{R_j})\in E}$ between a character node $V_{C_i}$ and a relation node $V_{R_j}$ when ${V_{C_i} = S_k}$ and ${V_{R_j} = R_k}$ in the triplet $T_k = (S_k, R_k, O_k)$.
\item[$\bullet$] ${e_{R_i,C_j}=(V_{R_i},V_{C_j})\in E}$ between a relation node $V_{R_i}$ and a character node $V_{C_j}$ when ${V_{R_i} = R_k}$ and ${V_{C_j} = O_k}$ in the triplet ${T_k = (S_k, R_k, O_k)}$.
\item[$\bullet$] ${e_{O_i,R_j}=(V_{O_i},V_{R_j})\in E}$ between an object node $V_{O_i}$ and a relation node $V_{R_j}$ when ${V_{O_i} = S_k}$ and ${{V_{R_j} = R_k}}$ in the triplet ${T_k = (S_k, R_k, O_k)}$.
\item[$\bullet$] ${e_{R_i,O_j}=(V_{R_i},V_{O_j})\in E}$ between a relation node $V_{R_i}$ and an object node $V_{C_j}$ when ${V_{R_i} = R_k}$ and ${V_{C_j} = O_k}$ in the triplet ${T_k = (S_k, R_k, O_k)}$.
\end{itemize}%
with $i$, $j$, and $k$ being the index for a certain object in a set. 

\subsection{Scene Description Generation}
Scene descriptions are generated from the video scene graph according to the set of rules in Table~\ref{tab:description} in an unsupervised manner. For each true condition in Table~\ref{tab:description}, a single sentence is generated. The final scene description is the concatenation of all the generated sentences, which serves as a representation of the semantic content in the video scene. Examples are shown in Fig.~\ref{fig:description}.

\begin{table}[t]
\caption{Sentence generation from the video scene graph, $G=(V, E)$ with $V = \{V_C, V_P, V_O, V_R, V_A\}$. We define $e_{R_k,O} = \{e_{R_k,O_j}\}$ with $j \in [1,...,|V_O|]$. $e_{R_k,C}$, $e_{O, R_k}$, and $e_{C, R_k}$ are defined likewise.}
\centering
\setlength{\tabcolsep}{3pt}
\resizebox{\textwidth}{!}{
\begin{tabular}{l l l}
\hline
\small \textbf{Graph Condition} & \small \textbf{Generated Sentence} & \small \textbf{Example}  \\
\hline
\small $|V_C| = 0$ \& $|V_P| = 0$ & \footnotesize \texttt{Someone is} \textcolor{black}{$~V_A$}. & \footnotesize Someone is lying on the floor.\\
\small $|V_C| = 1$ \& $|V_P| = 0$ & \footnotesize \textcolor{black}{$V_C~$} \texttt{is} \textcolor{black}{$~V_A$}. & \footnotesize Leonard is smiling.\\
$|V_C| > 1$ \& $|V_P| = 0$ & \footnotesize \textcolor{black}{$V_{C_1}, ..., V_{C_{|V_C|-1}}~$} \texttt{and} \textcolor{black}{$~V_{C_{|V_C|}}~$} \texttt{are} \textcolor{black}{$~V_A$}. & \footnotesize Penny and Amy are holding a bag.\\
\small $|V_C| = 0$ \& $|V_P| = 1$ & \footnotesize \texttt{Someone is} \textcolor{black}{$~V_A~$} \texttt{at} \textcolor{black}{$~V_P$}. & \footnotesize Someone is walking at the street.\\
\small $|V_C| = 1$ \& $|V_P| = 1$ & \footnotesize \textcolor{black}{$V_C~$} \texttt{is} \textcolor{black}{$~V_A~$} \texttt{at} \textcolor{black}{$~V_P$}. & \footnotesize  Sheldon is smiling at the bedroom. \\
\small $|V_C| > 1$ \& $|V_P| = 1$ & \footnotesize \textcolor{black}{$V_{C_1}, ..., V_{C_{|V_C|-1}}~$} \texttt{and} \textcolor{black}{$~V_{C_{|V_C|}}~$} \texttt{are} \textcolor{black}{$~V_A$} \texttt{at} \textcolor{black}{$~V_P$}. & \footnotesize  Amy and Raj are talking at the room.\\

\footnotesize $e_{C_i,R_k} \in E$ \& $e_{R_k,O_j} \in E$ \& $|e_{R_k,O}| = 1$ & \footnotesize \textcolor{black}{$V_{C_i} ~ V_{R_k} ~ V_{O_j}$}. & \footnotesize Penny wearing shorts.\\

\footnotesize $e_{C_i,R_k} \in E$ \& $e_{R_k,O_j} \in E$ \& $|e_{R_k,O}| > 1$ & \footnotesize \textcolor{black}{$V_{C_i} ~ V_{R_k} ~ V_{O_1}, ..., V_{O_{|V_O|-1}}~$} \texttt{and} \textcolor{black}{$~V_{O_{|V_O|}}$}. & \footnotesize Raj holding bottle and book. \\

\footnotesize $e_{R_k,C_i} \in E$ \& $e_{O_j,R_k} \in E$ \& $|e_{O,R_k}| = 1$ & \footnotesize \textcolor{black}{$V_{O_j} ~ V_{R_k} ~ V_{C_i}$}. & \footnotesize Board behind Sheldon. \\

\footnotesize $e_{R_k,C_i} \in E$ \& $e_{O_j,R_k} \in E$ \& $|e_{O,R_k}| > 1$ & \footnotesize \textcolor{black}{$V_{O_1}, ..., V_{O_{|V_O|-1}}~$} \texttt{and} \textcolor{black}{$~V_{O_{|V_O|}} ~ V_{R_k} ~ V_{C_i}$}. & \footnotesize  Chair, table and door behind Penny.\\
\hline
\end{tabular}}
\label{tab:description}
\end{table}

\subsection{Observing Transformer}
The generated description, $d$, is used in the input string for the observe branch: %
\begin{equation}
    {s_o}^{c} = \text{[CLS]} + d + q + \text{[SEP]} + a^c + \text{[SEP]},
\end{equation} %
Each ${s_o}^c$ is fed into the Observing Transformer to obtain ${\vec{y}_o}^c$, which is forwarded into a single output linear layer to compute the \textit{observe score}: %
\begin{equation}
{\alpha_o}^c = {\vec{w}_o}^\top \cdot {\vec{y}_{o}}^c + b_o
\end{equation}


\section{Recall Branch}
In the recall branch, ROLL emulates the human experience of watching a TV show by recalling the events that occurred previously. This is inspired by the human evaluation on \cite{garcia2020knowit}, which provides some insights on human behaviour. In \cite{garcia2020knowit}, evaluators were asked to answer questions about a popular sitcom under different conditions. Interestingly, the reported performance dropped dramatically when humans were not exposed to the videos. We speculate that this is because humans indirectly used the scene to remember the whole episode and answer questions about the plot. The recall branch imitates this behaviour by first identifying the video and then acquiring knowledge about the story plot.

\subsection{Knowledge Acquisition} 
Differently from previous work~\cite{garcia2020knowit}, in which the external knowledge to answer each question is specifically annotated by humans, we rely on publicly available resources\footnote{For example, \href{https://the-big-bang-theory.com/}{https://the-big-bang-theory.com/}} and build a knowledge base (KB) using plot summaries from the Internet.\footnote{Generating video plot summaries automatically from the whole video story is a challenging task by itself and out of the scope of this work. However, it is an interesting problem that we aim to study as a our future work.} Given a video scene, we first identify the video story it belongs to as in video retrieval \cite{Garcia2018Temporal}. Frames are represented by the output of the second-to-last layer of a pre-trained ResNet50~\cite{he2016deep}. We compute the cosine similarity between each frame representation in the scene and all frames in the dataset, keeping the video of the most similar frame. As a result, we obtain an identifier of the most voted video, which is used to query the KB and we obtain a document $p$  with the plot. In this way, ROLL acquires external knowledge about the video story in an weakly supervised way as 1) the questions and the external knowledge base have not been paired in any way during their generation, 2) the model does not know if there is corresponding text in the external knowledge base that can be useful for a given question, 3) the model is not directly trained with ground-truth episode labels, and 4) the model is not trained with ground-truth text location.

\subsection{Recalling Transformer}
The document $p$ is fed into the Recalling Transformer to predict a \textit{recall score} for each candidate answer. As many documents exceed the maximum number of words the Transformer can take as input,\footnote{In The Big Bang Theory, the longest  summary contains 1,605 words.} we adopt a sliding window approach \cite{hewlett2017accurate,hu2019retrieve} to slice $p$ into multiple overlapping segments. To produce the segments $k_j$ with $j = 1,\cdots,N_{s_{\text{MAX}}}$, we slide a window of length $W_l$ with a stride $r$ over the document $p$, obtaining $N_s = \lceil\frac{L_d - W_l}{r}\rceil + 1$ segments, where $L_d$ is the number of words in the document. For training multiple samples in a minibatch, we set all the documents to have the same number of segments $N_{s_{\text{MAX}}}$, discarding segments if $N_s > N_{s_{\text{MAX}}}$, and zero-padding if $N_s < N_{s_{\text{MAX}}}$. We encode the plot segments along with the question and candidate answers into multiple input strings: %
\begin{equation}
 {s_{ll}}_j^{c} = \text{[CLS]} + q + \text{[SEP]} + a^c + k_j + \text{[SEP]}
\end{equation} %
Each ${s_{ll}}_j^{c}$ is fed into the Recalling Transformer to obtain ${\vec{y}_{ll}}^c_j$, which is forwarded into a single output linear layer to compute a score for an answer-segment pair: %
\begin{equation}
{\alpha_{ll}}^c_j = {\vec{w}_{ll}}^\top \cdot {\vec{y}_{ll}}^c_j + b_{ll}
\end{equation} %
Then, the final recall score for each of the candidate answers ${\alpha_{ll}}^c$ is: %
\begin{equation}
{\alpha_{ll}}^c = \max({\alpha_{ll}}^c_j) \quad \text{ with } j = 1, \cdots, N_{s_{\text{MAX}}}
\end{equation}


\section{Final Prediction}
To output the final prediction score, the model concatenates the output of the three branches into a score vector $\boldsymbol{\alpha}^c = [{\alpha_{r}}^c, {\alpha_{o}}^c, {\alpha_{ll}}^c]$, which is input into a single layer classifier. The predicted answer $\hat{a}$ is then: %
\begin{equation}
\omega^c = {\vec{w}_{c}}^\top \cdot \boldsymbol{\alpha}^c + b_{c}
\end{equation}%
\begin{equation}
\hat{a} = a^{\argmax_{c} \boldsymbol{\omega}} \quad \text{with } \boldsymbol{\omega} = [{\omega}^1, \cdots, {\omega}^{N_{ca}}]^\top
\end{equation}

\noindent
\subsubsection{Modality Weighting Mechanism}
Wang et al. \cite{wang2019makes} have shown that multi-modality training often suffers from information loss, degradating performance with respect to single modality models. To avoid losing information when merging the three branches in ROLL, we use a modality weighting (MW) mechanism. First, we ensure that each Transformer learns independent representations by training them independently. The multi-class cross-entropy loss is computed as:%
\begin{equation}
\mathcal{L}(\boldsymbol{\delta}, c^*) =  - \log \frac{\exp({\delta}^{c^*})}{\sum_c \exp({\delta}^c)} 
\end{equation} %
where $c^*$ is the correct answer, and $\boldsymbol{\delta} = [{\delta}^1, \cdots, {\delta}^{N_{ca}}]$ the vector with the scores of the candidate answers. Next, the Transformers are frozen and the three branches are fine-tuned together. To ensure the multi-modal information is not lost, the model is trained as a multi-task problem with $\beta_r + \beta_o + \beta_{ll} + \beta_{\omega} = 1$: %
\begin{equation*}
\small
\mathcal{L}_{\text{MW}} =  \beta_r \mathcal{L}(\boldsymbol{\alpha}_r, c^*) + \beta_o \mathcal{L}(\boldsymbol{\alpha}_o, c^*) +\beta_{ll} \mathcal{L}(\boldsymbol{\alpha}_{ll}, c^*) + \beta_{\omega} \mathcal{L}(\boldsymbol{\omega}, c^*)
\end{equation*} %

\section{Evaluation}

\noindent
\textbf{Datasets}
We evaluate ROLL on the KnowIT VQA~\cite{garcia2020knowit} and the TVQA+~\cite{lei2019tvqa+} datasets. KnowIT VQA is the only dataset for knowledge-based video story question answering, containing 24,282 questions about 207 episodes of The Big Bang Theory TV show. Questions in the test set are divided into four categories: visual-based, textual-based, temporal-based, and knowledge-based, and each question is provided with ${N_{ca} = 4}$ candidate answers. Accuracy is computed as the number of correct predicted answers over the total number of questions. Even though our model is specifically designed for leveraging external knowledge, we also evaluate its generalisation performance on non knowledge-based video story question answering. For this purpose, we use the TVQA+ dataset, in which questions are compositional and none of them requires external knowledge. TVQA+ contains 29,383 questions, each with ${N_{ca} = 5}$ candidate answers.

\begin{table}[t]
\centering
\caption{Evaluation on KnowIT VQA test set.}
\resizebox{0.98\textwidth}{!}{
\begin{tabular}{l l l l l l l c c c c c}
\hline
 & & & \multicolumn{3}{c}{\textbf{Data}} & & \multicolumn{5}{c}{\textbf{Accuracy}}\\
 \cline{4-6}
  \cline{8-12}
\textbf{Method} & \textbf{Encoder} & & \textbf{Dialog} & \textbf{Vision} & \textbf{Know.} & & \textbf{Vis.} & \textbf{Text.} & \textbf{Temp.} & \textbf{Know.} & \textbf{All}\\
\hline
Rookies \cite{garcia2020knowit} & - & & - & - & No & & 0.936 & 0.932 & 0.624 & 0.655 & 0.748 \\
Masters \cite{garcia2020knowit} & - & & - & - & Yes & & 0.961 & 0.936 & 0.857 & 0.867 & 0.896 \\
\hline
TVQA \cite{lei2018tvqa} & LSTM & & Subs. & Concepts & - & & 0.612 & 0.645 & 0.547 & 0.466 & 0.522 \\
ROCK {\scriptsize Img} \cite{garcia2020knowit} & BERT & & Subs. & ResNet & Human & & 0.654 & 0.681 & 0.628 & 0.647 & 0.652 \\
ROCK {\scriptsize Cpts} \cite{garcia2020knowit} & BERT & & Subs. & Concepts & Human & & 0.654 & 0.685 & 0.628 & 0.646 & 0.652 \\
ROCK {\scriptsize Faces} \cite{garcia2020knowit} & BERT & & Subs. & Characters & Human & & 0.654 & 0.688 & 0.628 & 0.646 & 0.652\\
ROCK {\scriptsize Caps} \cite{garcia2020knowit} & BERT & & Subs. & Captions & Human & & 0.647 & 0.678 & 0.593 & 0.643 & 0.646 \\
\hline
ROLL-human & BERT & & Subs. & Descriptions & Human & & 0.708 &	\textbf{0.754} & 0.570 &	0.567	& 0.620	 \\
ROLL & BERT & & Subs. & Descriptions & Summaries & & \textbf{0.718} & 0.739 & \textbf{0.640} & \textbf{0.713} & \textbf{0.715}\\
\hline
\end{tabular}
}
\label{tab:sota}
\end{table}

\noindent
\textbf{Implementation Details} We use the BERT uncased base model with pre-trained initialisation for our three Transformers. The maximum number of tokens is set to 512. For the single branch training, transformers are fine-tuned following the details in~\cite{devlin2019bert}. For the joint model training, we use stochastic gradient descent with momentum 0.9 and learning rate 0.001. In the observe branch, we extract the frames for the Character Recognition, Place Classification and Object Relation Detection modules at 1 fps, and for the Action Recognition module at 24 fps. In total, we use 17 characters, 32 places, 150 objects, 50 relations, and 157 action categories. In the recall branch, we use a window length $W_l = 200$, stride $r = 100$, and maximum number of segments $N_{s_{\text{MAX}}} = 5$. In the modality weighting mechanism, we set $\beta_r = 0.06$, $\beta_o = 0.06$, $\beta_{ll} = 0.08$, and $\beta_{\omega} = 0.80$ unless otherwise stated. 

\noindent
\textbf{Evaluation on KnowIT VQA}
We compare ROLL against the latest reported results on the KnowIT VQA dataset: TVQA and  four different models in ROCK. TVQA~\cite{lei2018tvqa} is based on a two-stream LSTM encoder for subtitles and visual concepts, whereas ROCK~\cite{garcia2020knowit} uses task-specific human annotations to inform a BERT based model with external knowledge. ROCK reports results using four different visual representations: ResNet features, visual concepts, list of characters, and generated captions. For a more complete comparison, we also report results of ROLL using the human annotations from \cite{garcia2020knowit} as external knowledge (ROLL-human). Results are found in Table~\ref{tab:sota}. Main findings are summarised as:
\begin{enumerate}
    \item Overall, ROLL outperforms previous methods in all the question categories by a large margin, with 6.3\% improvement on the overall accuracy with respect to the best performing ROCK.
    \item When comparing the visual representations, our proposed video descriptions contain more semantic information than previous methods, improving visual-based questions by at least 6.4\%. Specially, the boost in performance in visual-based questions with respect to standard captioning (ROCK {\footnotesize Captions}) or visual concepts (ROCK {\footnotesize Concepts}, TVQA) validates our unsupervised video descriptions as the best representation for this task.
    \item Additional evidence of the superior performance of our proposed unsupervised descriptions is shown when ROLL uses human annotations as external knowledge. Although the overall performance is lower because ROLL-human is not optimised to exploit this kind of information, on the visual-based questions our method improves best previous work by 5.4\%. As  the same source of knowledge is used, the superior performance can only be due to the contribution of our proposed visual representations. 
    \item On the knowledge-based samples, our method based on plot summaries outperforms task-specific human annotations by 6.7\%, even when less annotations are required. This implies that the proposed slicing mechanism in the recall branch successfully extracts the relevant information from the long documents provided as external knowledge.
    \item When compared against human performance, ROLL is 18\% behind masters accuracy (humans that have watched the show) and it is closer to rookies non-knowledge accuracy (humans that have never watched the show). This shows how challenging this task is, with still plenty room for improvement.
\end{enumerate}

\begin{figure}[t]
  \begin{minipage}{\textwidth}
    \begin{minipage}[b]{0.48\textwidth}
    \centering
    \renewcommand{\arraystretch}{1.1}
    \small
    \setlength{\tabcolsep}{3pt}
    \resizebox{0.98\textwidth}{!}{
    \begin{tabular}{l l l r}
    \hline
    \textbf{Method} & \textbf{Vision} & \textbf{Lang.} & \textbf{Acc}\\
    \hline
    TVQA \cite{lei2018tvqa} & Concepts & LSTM & 62.28 \\
    TVQA \cite{lei2018tvqa} & Regional & LSTM & 62.25 \\
    STAGE \cite{lei2019tvqa+} & Regional & GloVe & 67.29 \\
    STAGE \cite{lei2019tvqa+} & Regional & BERT & 68.31 \\
    ROLL & Description & BERT & \textbf{69.61} \\
    \hline
    \end{tabular}}
    \captionof{table}{Evaluation on the TVQA+ val set. No external knowledge is used.}\label{tab:tvqa}
    \end{minipage}
  \hfill
    \begin{minipage}[b]{0.48\textwidth}
    \centering
    \includegraphics[width = 0.95\textwidth]{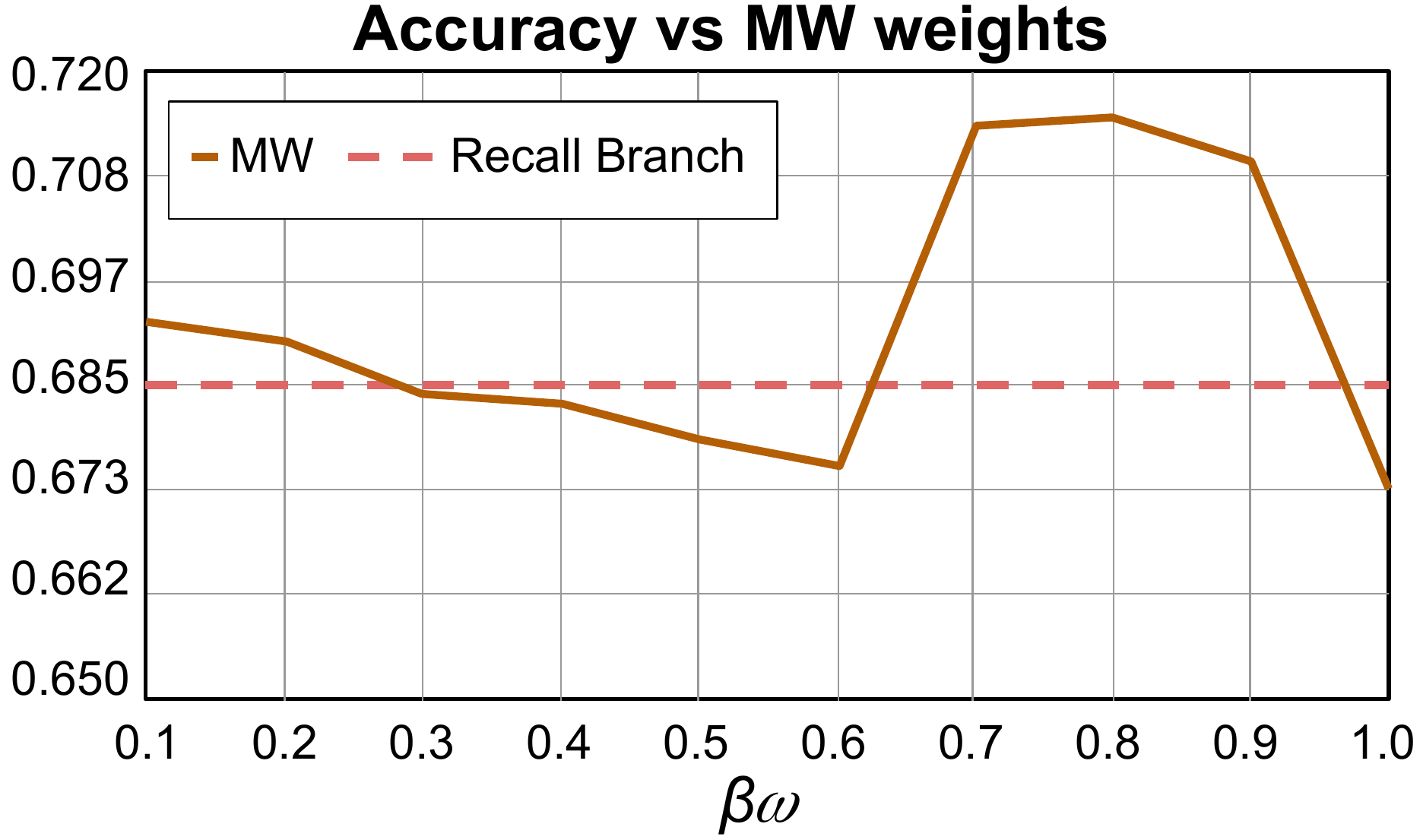}
    \vspace{-10pt}
  \captionof{figure}{ROLL accuracy according to $\beta_{\omega}$ on KnowIT VQA test set.}\label{fig:MW}
  \end{minipage}
  \end{minipage}
\end{figure}

\noindent
\textbf{Evaluation on TVQA+}
To show ROLL generalisation performance even when external knowledge is not necessary, we additionally evaluate it on the TVQA+ dataset. For a fair comparison against previous work, 1) we remove the recall branch in ROLL and only use the read and observe branches, i.e., no external knowledge is used, and 2) we compare ROLL against models that use the answer labels as the only supervision for training, i.e., no extra annotations such as timestamps or spatial bounding boxes are used. Results are found in Table~\ref{tab:tvqa}. Consistent with the results on the KnowIT VQA dataset, ROLL based on scene descriptions outperforms models based on other visual representations, such as visual concepts or Faster R-CNN \cite{ren2015faster} regional features, by at least 1.3\%.

\noindent
\textbf{Ablation study}
We perform an ablation study to measure the contribution of each branch. Results when using one, two, or the three branches on the KnowIT VQA dataset are reported in Table~\ref{tab:branches}. When a single branch is used, the observe branch gets the worst overall accuracy and the recall branch performs the best. This is consistent with the types of questions in the dataset, with 22\% being visual-based and 63\% being knowledge-based. The read branch gets the best performance in the text-based questions (i.e., about the subtitles), and the recall branch gets the best accuracy in the knwoledge-based questions (i.e., about the storyline). When the observe branch is combined with other branches it consistently contributes to improve the results. Again, this result strongly suggests that the generated scene descriptions do contain meaningful information for the task. The full model combining the three branches performs the best.

\begin{table}[t]
\parbox{.49\linewidth}{
\centering
\caption{ROLL ablation study.}
\label{tab:branches}
\resizebox{0.49\textwidth}{!}{
\begin{tabular}{l c c c c c c}
\hline
\textbf{Branch} &  \textbf{Vis.} & \textbf{Text.} & \textbf{Temp.} & \textbf{Know.} & \textbf{All}\\
\hline
{\scriptsize Read}   & 0.656 & \textbf{0.772} & 0.570 & 0.525 & 0.584 \\
{\scriptsize Observe}& 0.629 & 0.424 & 0.558 & 0.514 & 0.530 \\
{\scriptsize Recall} & 0.624 & 0.620 & 0.570 & \textbf{0.725} & 0.685 \\
{\scriptsize Read-Observe} & 0.695	& 0.732 & 0.570 & 0.527	& 0.590\\
{\scriptsize Observe-Recall} & 0.712 &	0.601 &	0.628 &	0.704 &	0.691 \\
{\scriptsize Read-Recall} & \textbf{0.722} & 0.732 & 0.628 & 0.708 & 0.711\\
{\scriptsize Full Model} & 0.718 & 0.739 & \textbf{0.640} & 0.713 & \textbf{0.715}\\
\hline
\end{tabular}}
}
\hfill
\parbox{.49\linewidth}{
\centering
\caption{Fusion Methods Comparison.}
 \label{tab:fusion}
\resizebox{0.49\textwidth}{!}{
\begin{tabular}{l c c c c c}
\hline
\textbf{Method} & \textbf{Vis.} & \textbf{Text.} & \textbf{Temp.} & \textbf{Know.} & \textbf{All}\\
\hline
Average & 0.726 & 0.710 & 0.628	& 0.648	& 0.672 \\
Maximum & 0.685	& 0.757	& 0.593	& 0.678	& 0.686 \\
Self-att  & \textbf{0.737} &	\textbf{0.761} & \textbf{0.651} & 0.641 &	0.677 \\
QA-att & 0.736 & 0.743 & 0.605 & 0.637 & 0.670\\
FC w/o MW & 0.728 & 0.743 & 0.616 & 0.637 &	0.669\\
FC w/ MW  & 0.718 & 0.739 & 0.640 & \textbf{0.713} & \textbf{0.715}\\
\hline
\end{tabular}}
}
\end{table}

\noindent
\textbf{Fusion Methods Comparison}
We also study the performance of our proposed MW mechanism and compare it against several fusion methods. Results are reported in Table~\ref{tab:fusion}. Given the three prediction scores from each of the branches, Average and Maximum compute the average and maximum score, respectively. The Self-att method implements a self-attention mechanism based on the Transformer outputs, and the QA-att mechanism attends each of the modality predictions based on the BERT representation of the question and candidate answers. The  FC w/o MW predicts the answer scores by concatenating the scores of the three branches and feeding them into a linear layer, and FC w/ MW builds our proposed MW mechanism on top. The results show that most of the methods fail at properly fusing the information from the three branches, i.e., the overall performance is lower than the best single branch (recall, as reported in Table~\ref{tab:branches}). This is probably because the fusion of the different modalities incurs in information loss. Our MW mechanism, in contrast, successfully balances the contribution from the three branches. Fig.~\ref{fig:MW} compares different values of $\beta_{\omega}$ in the MW against the best performing single modality, with $\beta_r$, $\beta_o$, and $\beta_{ll}$ uniformly distributed. When the MW is not used ($\beta_{\omega} = 1$) the model obtains the worst performance. Likewise, when the loss contribution from the final prediction is too weak ($\beta_{\omega} < 0.6$), the model is not able to fuse the information correctly.

\begin{figure*}[t]
\centering
\begin{tabular}{c}
\includegraphics[width = 0.95\textwidth]{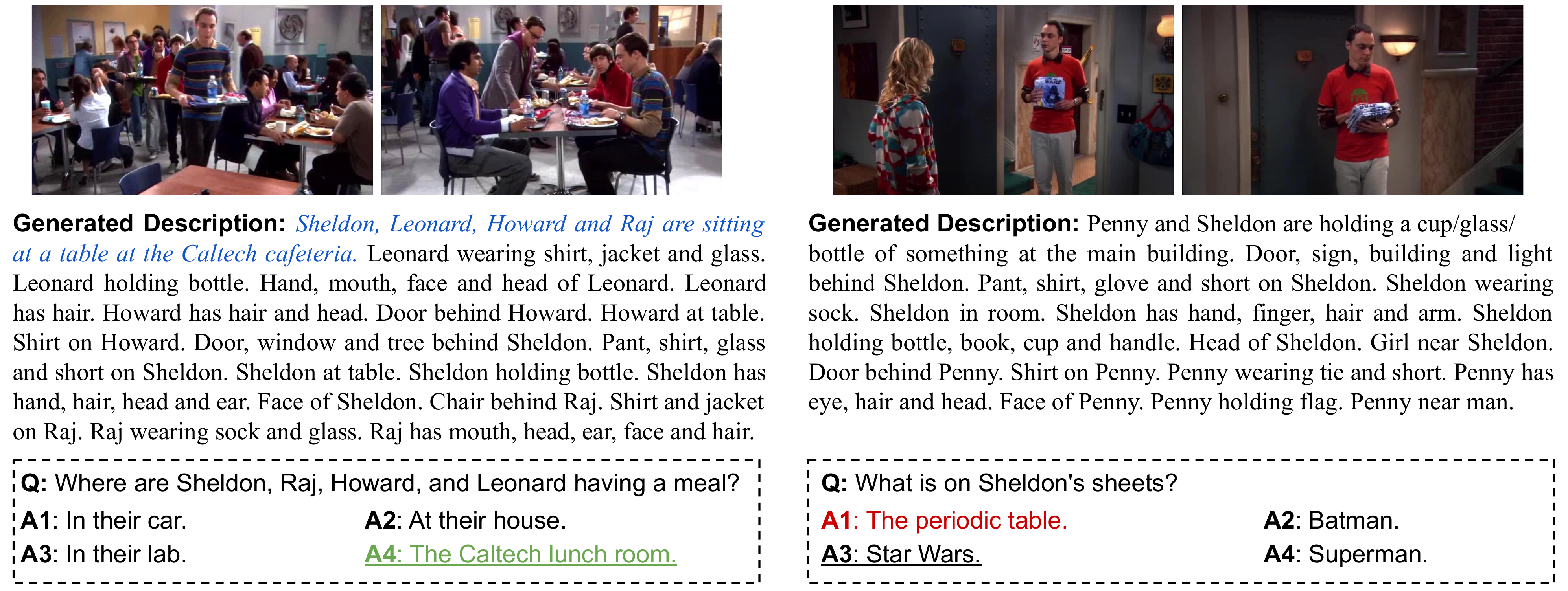}
\end{tabular}
\caption{ROLL visual results. Underline/colour for correct/predicted answers. The relevant part for the question in the generated description is highlighted in blue.}
\label{fig:results}
\end{figure*}

\begin{figure}[t]
\centering
\begin{tabular}{c}
\includegraphics[width = 0.95\textwidth]{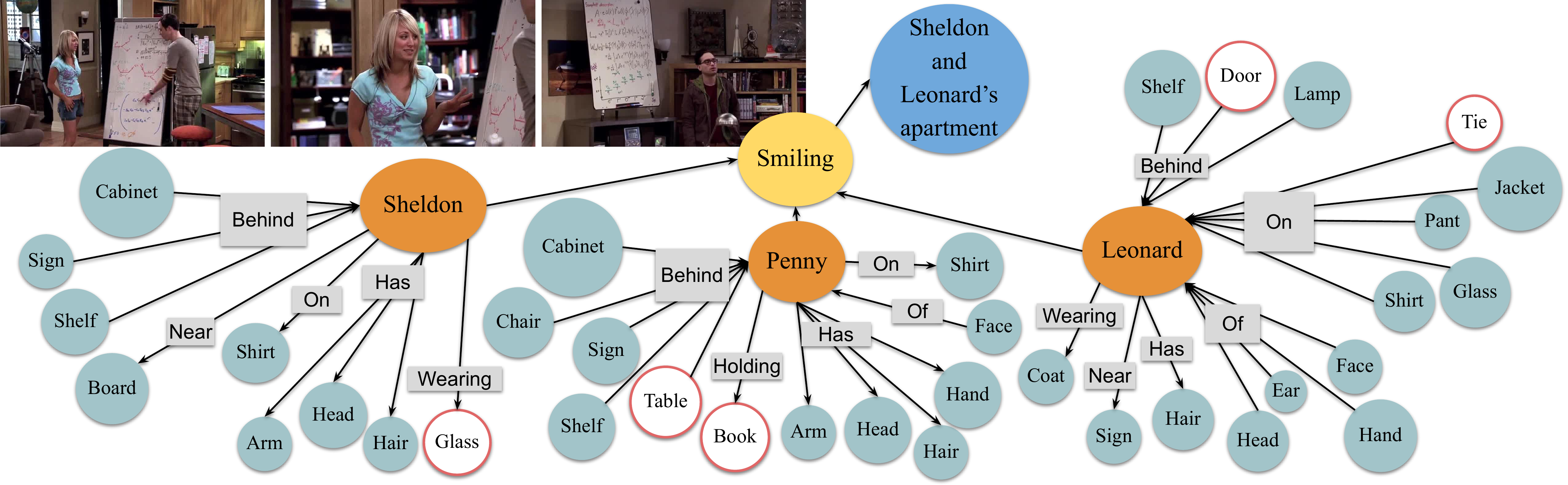}
\end{tabular}
\caption{Generated scene graph. Solid for correct and bordered for incorrect nodes.}
\label{fig:graph}
\end{figure}

\noindent
\textbf{Qualitative Results}
We visually inspect ROLL results to understand the strengths and weaknesses of our model. An example of scene graph can be seen in Fig.~\ref{fig:graph}, whereas results on visual-based questions are provided in Fig.~\ref{fig:results}. ROLL performs well on questions related to the general content of the scene, such as places, people, or objects, but fails in detecting fine-grained details. Performance of the individual video modules is reported in the supplementary material.

\section{Conclusion}

We introduced ROLL, a model for knowledge-based video story question answering. To extract the visual information from videos, ROLL generates video descriptions in an unsupervised way by relying on video scene graphs. This new video representation led the model to an important increase of accuracy on visual-based questions on two datasets. Moreover, unlike previous work, ROLL leverages information from external knowledge without specific annotations on the task, easing the requirements of human labelling. This came without a drop in performance. On the contrary, as ROLL successfully fuses specific details from the scene with general information about the plot, the accuracy in KnowIT VQA and TVQA+ datasets was improved by more than 6.3\% and 1.3\%, respectively. Finally, by incorporating a modality weighting mechanism, ROLL avoided the information loss that comes from fusing different sources.

\noindent
\textbf{Acknowledgement} This work was supported by a project commissioned by the New Energy and Industrial Technology Development Organization (NEDO), and JSPS KAKENHI Nos.~18H03264 and 20K19822. We also would like to thank the anonymous reviewers for they insightful comments to improve the paper.

\clearpage

\bibliographystyle{splncs04}
\bibliography{eccv}

\clearpage
\appendix
\section{Spatio-Temporal Character Filter}

To remove mispredictions and duplicate detections in the character recognition module, we apply a spatio-temporal filter by finding all the faces belonging to the same person and re-assigning character names based on temporal coherence. For each scene, we first detect frames belonging to the same shot (i.e., captured with the same camera) by computing the cosine similarity between their ResNet50 representations, $\vec{f}_{\text{res}} \in \mathbb{R}^{2048}$. Then, for all the frames in a shot, we compute the Euclidean distance between all possible pairs of faces using their bounding box centroids. When a pair of faces is close (i.e., the distance is below a threshold\footnote{Experimentally set to 50.}), faces are considered to be the same. For all the faces assigned to the same person, we find the most frequent name predicted by the kNN classifier. If the name appears in at least 70\% of the frames in the shot, the character is assigned to that name. Otherwise, the character is assigned as \textit{unknown}. 

\section{Fusion Methods Details}
Given the output scores for each branch, ${\alpha_{r}}^c$, ${\alpha_{o}}^c$, and ${\alpha_{ll}}^c$, the final score for each candidate answer $\omega^c$ is computed according to each fusion method as:

\begin{itemize}
    \item[$\bullet$] \textit{Average}: $~\omega^c = \frac{1}{3} ({\alpha_{r}}^c + {\alpha_{o}}^c + {\alpha_{ll}}^c)$
    
    \item[$\bullet$] \textit{Maximum}: $~\omega^c = \max({\alpha_{r}}^c, {\alpha_{o}}^c, {\alpha_{ll}}^c)$
    
    \item[$\bullet$] \textit{Self-att}: According to the Transformers embeddings ${\vec{y}_{r}}^c$, ${\vec{y}_{o}}^c$, and ${\vec{y}_{ll}}^c$, with ${{\vec{y}_{ll}}^c = {\vec{y}_{ll}}^c_{j_{\text{max}}}}$ and ${j_{\text{max}} = \argmax_j({\alpha_{ll}}^c_j)}$, the self-attention weights are computed as: 
    \begin{equation*}
    \small
        {\boldsymbol{\psi}}^c = {\vec{W}_{\text{self}}} \begin{pmatrix}{\vec{y}_{r}}^c \\ {{\vec{y}_{o}}^c} \\ {{\vec{y}_{ll}}^c}
        \end{pmatrix} + \vec{b}_{\text{self}}
    \end{equation*} %
    with ${{\boldsymbol{\psi}}^c = [{\psi_r}^c, {\psi_o}^c, {\psi_{ll}}^c]}$. The information from the three branches is fussed according to the self-attention weights as ${\vec{y}_{\text{self}}^c = {\psi_r}^c{\vec{y}_{r}}^c + {\psi_o}^c{\vec{y}_{o}}^c + {\psi_{ll}}^c{\vec{y}_{ll}}^c}$, and the final score for each candidate answer is ${\omega^c = {\vec{w}_{c}}^\top \cdot {\vec{y}_{\text{self}}}^c + b_{c}} $.
    
    \item[$\bullet$] \textit{QA-att}: According to the output of the pre-trained BERT network when using the question and a candidate answer, ${\vec{y}_{\text{QA}}}^c$, the attention weight for the read branch is computed as:
    \begin{equation*}
    \small
        {\psi_r}^c = {\vec{w}_{\text{att}}}^\top \cdot \begin{pmatrix}{\vec{y}_{r}}^c \\ {{\vec{y}_{QA}}^c}
        \end{pmatrix} + b_{\text{att}}
    \end{equation*} %
    with ${\psi_o}^c$ and ${\psi_{ll}}^c$ computed equivalently. The information from the three branches is fused according to the attention weights as $\vec{y}_{\text{att}}^c = {\psi_r}^c{\vec{y}_{r}}^c + {\psi_o}^c{\vec{y}_{o}}^c + {\psi_{ll}}^c{\vec{y}_{ll}}^c$, and the final score for each candidate answer is computed as ${\omega^c = {\vec{w}_{c}}^\top \cdot {\vec{y}_{\text{att}}}^c + b_{c}} $.
    
    \item[$\bullet$] \textit{Linear w/o MW}: A linear layer trained without MW mechanism fuses the three branches information, ${\omega^c = {\vec{w}_{c}}^\top \cdot [{\alpha_{r}}^c, {\alpha_{o}}^c, {\alpha_{ll}}^c]^\top + b_{c}}$ and $\beta_{\omega} = 1$.
    \item[$\bullet$] \textit{Linear w/ MW} A linear layer trained with our MW mechanism fuses the three branches information, ${\omega^c = {\vec{w}_{c}}^\top \cdot [{\alpha_{r}}^c, {\alpha_{o}}^c, {\alpha_{ll}}^c]^\top + b_{c}}$ and $\beta_{\omega} < 1$.
\end{itemize}

\section{Video Modules Accuracy}
\begin{figure}
\centering
\begin{tabular}{cc}
\includegraphics[width = 0.48\textwidth]{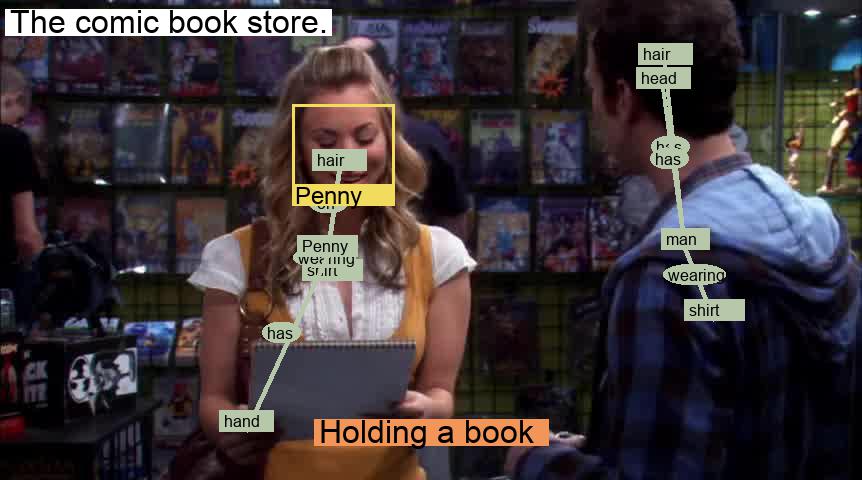} &
\includegraphics[width = 0.48\textwidth]{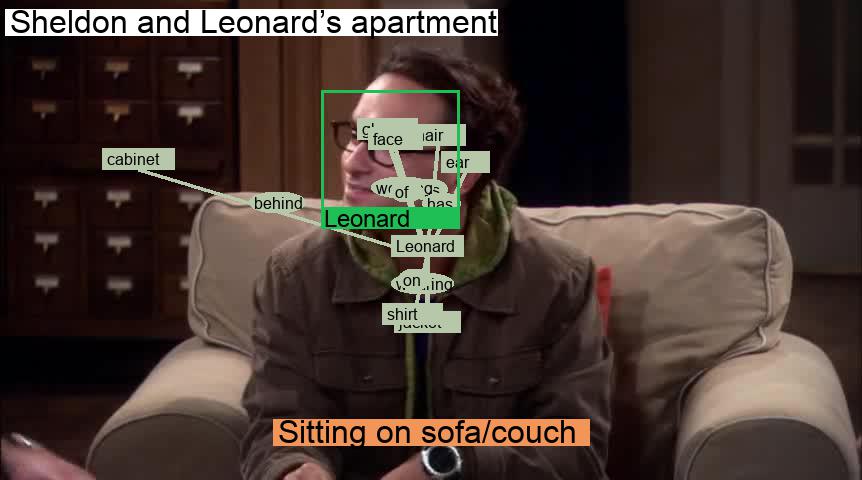} \\
\includegraphics[width = 0.48\textwidth]{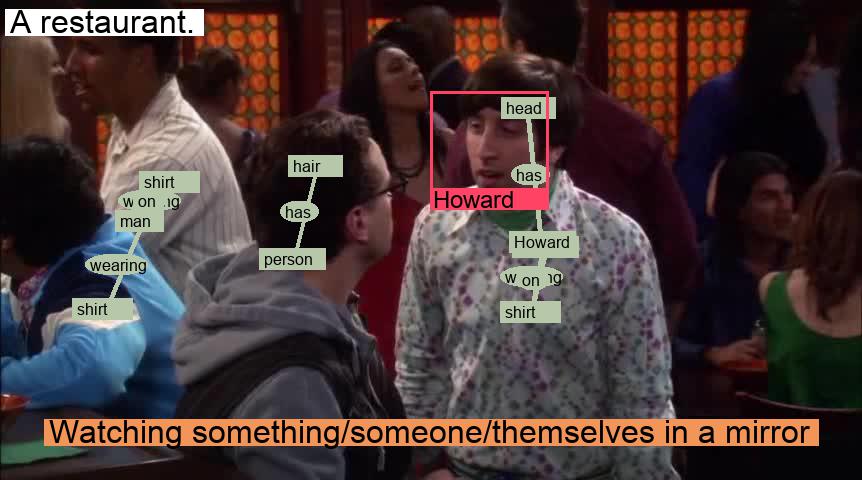} &
\includegraphics[width = 0.48\textwidth]{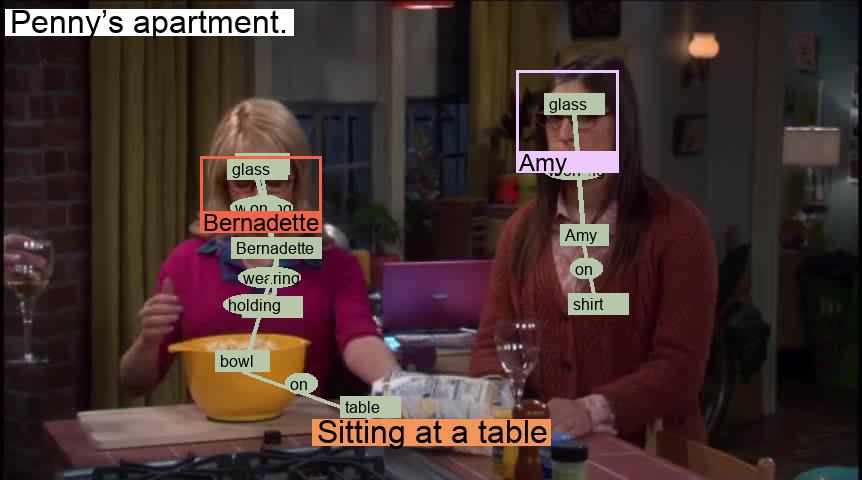} \\
\includegraphics[width = 0.48\textwidth]{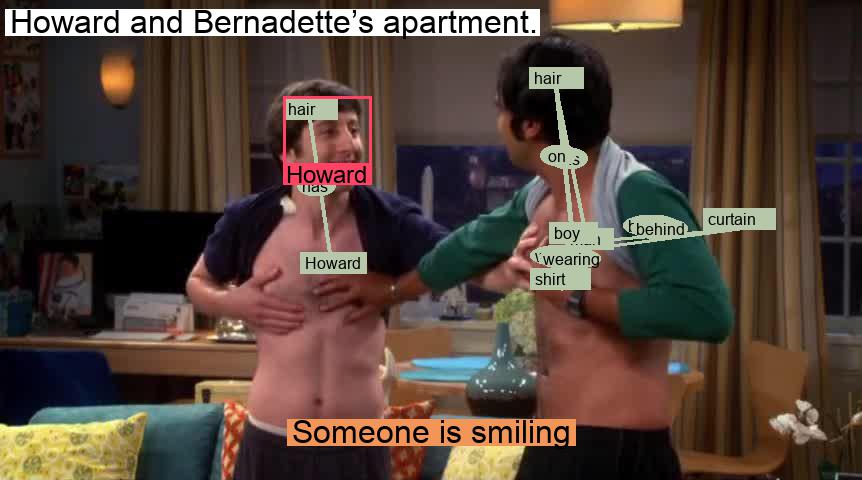} &
\includegraphics[width = 0.48\textwidth]{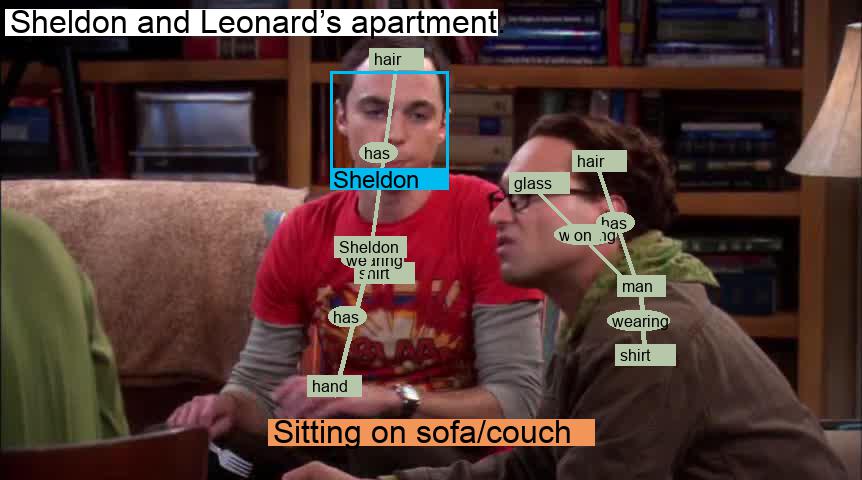} \\
\end{tabular}
\caption{Predictions of the video modules. For each frame, faces and objects on their bounding boxes, location on the top right, and action on the bottom middle.}
\label{fig:modules_examples}
\end{figure}

We evaluate the performance of the different video modules used in the observe branch. As datasets in video story question answering do not generally contain annotations on characters, places, objects relationships or actions, accuracy results are difficult to obtain. Even so, we report accuracy for the Character Recognition (Table \ref{tab:characteracc}) and the Place Classification (Table \ref{tab:placeracc}) modules. For the Character Recognition, we randomly annotate a subset of 100 video frames and measure accuracy as the number of correct predictions before and after applying the spatio-temporal filter. For the Place Classification, we report accuracy before and after the temporal filter on the KnowIT VQA test set according to the labels obtained from the scripts. Additionally, visual results with the predictions from the four modules (Character Recognition, Place Classification, Object Relation Detection, and Action Recognition) are in Fig. \ref{fig:modules_examples}. Finally, the list of characters and places can be found in Tables \ref{tab:characters} and \ref{tab:places}, respectively. 

\begin{table}
\parbox{.48\linewidth}{
\small
\centering
\caption{Character Recognition accuracy on 100 random frames.}
\begin{tabular}{l c c c}
\hline
 & \textbf{Precision} & \textbf{Recall} & \textbf{F1-score}\\
\hline
Before filter & 86.11 &	89.90 &	87.96 \\
After filter &  75.76 &	73.99 &	74.86 \\
\hline
\end{tabular}
\label{tab:characteracc}
}
\hfill
\parbox{.48\linewidth}{
\small
\centering
\caption{Place Classification accuracy on the KnowIT VQA test set.}
\begin{tabular}{l c c}
\hline
 & \textbf{Prec@1} & \textbf{Prec@5} \\
\hline
Before filter & 63.13  & 84.87 \\
After filter & 66.43  & - \\
\hline
\end{tabular}
\label{tab:placeracc}
}
\end{table}

\begin{table*}
\caption{List of characters.}
\footnotesize
\centering
\resizebox{0.97\textwidth}{!}{
\begin{tabular}{| l l l l l | }
\hline
Sheldon &
Leonard &
Penny &
Howard &
Raj \\
Amy &
Bernadette &
Dr. Beverly Hofstadter &
Stuart &
Barry \\
Emily &
Leslie &
Lucy &
Mary Cooper &
Priya \\
Dr. VM Koothrappali &
Wil Wheaton & & & \\
\hline
\end{tabular}}
\label{tab:characters}
\end{table*}

\begin{table*}
\caption{List of places.}
\footnotesize
\centering
\begin{tabular}{| l l | }
\hline
Penny's apartment &
The main building \\
Sheldon and Leonard's apartment &
Penny's apartment door \\
A lab &
A restaurant \\
A party &
A car \\
Sheldon's bedroom &
An office \\
The Cheesecake Factory &
A room \\
Leonard's bedroom &
Howard's bedroom \\
The cinema &
The Caltech cafeteria \\
Caltech University &
Sheldon's office \\
A store &
The hospital \\
Raj's apartment &
The laundry room \\
Penny's bedroom  &
The comic book store  \\
A house &
A bathroom  \\
Outside a house or a building &
A bar \\
Howard's house &
Amy's apartment \\
Howard and Bernadette's apartment &
Howard and Bernadette's house \\
\hline
\end{tabular}
\label{tab:places}
\end{table*}

\end{document}